\pdfoutput=1
\documentclass[11pt]{article}
\usepackage[preprint]{acl}
\usepackage[english]{babel}
\usepackage[T1]{fontenc}
\usepackage[utf8]{inputenc}
\usepackage{times,microtype,booktabs,graphicx,bm,amssymb,amsmath,bbding}
\usepackage{inconsolata,bigdelim,xspace,dsfont,fancyvrb,cleveref,multirow}
\makeatletter \@namedef{T1/zi4/m/it}{<->ssub*zi4/m/n} \makeatother 

\PassOptionsToPackage{frozencache=true,cachedir=minted-cache}{minted}
\usepackage[most]{tcolorbox}
\tcbuselibrary{minted}
\tcbuselibrary{xparse}
\usepackage{minted}

\tcbset{
    pythoncodebox/.style={
        enhanced jigsaw,breakable,
        colback=gray!10,colframe=gray!20!black,
        boxrule=1pt,top=2pt,bottom=2pt,left=2pt,right=2pt,
        sharp corners,before skip=10pt,after skip=10pt,
        attach boxed title to top left,
        boxed title style={empty,
            top=0pt,bottom=0pt,left=2pt,right=2pt,
            interior code={\fill[fill=black!70] (frame.south west)
                --([yshift=-4pt]frame.north west)
                to[out=90,in=180] ([xshift=4pt]frame.north west)
                --([xshift=-8pt]frame.north east)
                to[out=0,in=180] ([xshift=16pt]frame.south east)
                --cycle;
            }
        },
        title={#1}, 
        fonttitle=\sffamily\bfseries
    },
    pythoncodebox/.default={}, 
    pythoncodebox*/.style={
        enhanced jigsaw,breakable,
        colback=gray!10,coltitle=gray!20!black,colbacktitle=tcbcolback,
        frame hidden,
        top=2pt,bottom=2pt,left=2pt,right=2pt,
        sharp corners,before skip=10pt,after skip=10pt,
        attach boxed title to top text left={yshift=-1mm},
        boxed title style={empty,
            top=0pt,bottom=0pt,left=2pt,right=2pt,
            interior code={\fill[fill=tcbcolback] (interior.south west)
                --([yshift=-4pt]interior.north west)
                to[out=90,in=180] ([xshift=4pt]interior.north west)
                --([xshift=-8pt]interior.north east)
                to[out=0,in=180] ([xshift=16pt]interior.south east)
                --cycle;
            }
        },
        title={#1}, 
        fonttitle=\sffamily\bfseries
    },
    pythoncodebox*/.default={}, 
}

\NewTCBListing{python}{ !O{} !D(){} !G{} }{
    listing engine=minted,
    listing only,
    pythoncodebox={#1}, 
    minted language=python,
    minted options/.expanded={
        autogobble,breaklines,mathescape,
        #2 
    },
    #3 
}

\NewTCBListing{python*}{ !O{} !D(){} !G{} }{
    listing engine=minted,
    listing only,
    pythoncodebox*={#1}, 
    minted language=python,
    minted options/.expanded={
        autogobble,breaklines,mathescape,
        #2 
    },
    #3 
}

\newcommand{\py}[1]{\mintinline{python}{#1}\xspace}
\newcommand{\dataset}{\py{Dataset}}
\newcommand{\cell}{\py{Cell}}
\newcommand{\task}{\py{Task}}
\newcommand{\score}{\py{Score}}
\newcommand{\subsampler}{\py{Subsampler}}
\newcommand{\on}{\textsc{on}\xspace}
\newcommand{\by}{\textsc{by}\xspace}
\newcommand{\across}{\textsc{across}\xspace}
\newcommand{\onset}{t_\text{on}}
\newcommand{\offset}{t_\text{off}}

\newcommand{\emoji}[1]{\,\raisebox{-0.1em}{\includegraphics[height=1em]{emojis/#1.pdf}}}
\newcommand{\greencheckmark}{\textcolor{OliveGreen}{\CheckmarkBold}\xspace}
\newcommand{\redcross}{\textcolor{BrickRed}{\XSolidBold}\xspace}

\title{fastabx: A library for efficient computation of ABX discriminability}
\author{
    Maxime Poli, Emmanuel Chemla, Emmanuel Dupoux \\
  ENS - PSL, EHESS, CNRS \\
  \texttt{maxime.poli@ens.psl.eu}
}

\begin{document}

\maketitle

\begin{abstract}
We introduce fastabx, a high-performance Python library for building ABX discrimination tasks. ABX is a measure of the separation between generic categories of interest. It has been used extensively to evaluate phonetic discriminability in self-supervised speech representations. However, its broader adoption has been limited by the absence of adequate tools. fastabx addresses this gap by providing a framework capable of constructing any type of ABX task while delivering the efficiency necessary for rapid development cycles, both in task creation and in calculating distances between representations. We believe that fastabx will serve as a valuable resource for the broader representation learning community, enabling researchers to systematically investigate what information can be directly extracted from learned representations across several domains beyond speech processing. The source code is available at
\url{https://github.com/bootphon/fastabx}.
\end{abstract}

\section{Introduction}
Self-supervised learning (SSL) has revolutionized speech processing by enabling models to learn useful representations from unlabeled audio data, leading to notable improvements in various downstream tasks \citep{mohamed2022selfsupervised,baevski2020wav2vec,hsu2021hubert,yang2021superb}. To systematically evaluate these representations, `universal' benchmarks \citep{yang2021superb} were developed, providing standardized protocols for several speech tasks. In these benchmarks, evaluation relies on supervised probes to measure performance. Although such an approach is not entirely robust to the architecture of the probe itself \citep{ZAIEM2025101695}, this represents a significant paradigm shift. The success of even simple probes suggests that practitioners now find useful information to be readily accessible from the representations.

\begin{figure}[ht]
    \centering
    \begin{tikzpicture}[thick]
    \node[anchor=north west] (A) at (0,0) {$a = $\emoji{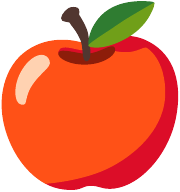}};
    \node[anchor=north west] (X) at (1,2) {$x = $\scalebox{1.4}{\emoji{red-apple}}};
    \node[anchor=north west] (B) at (5,0) {$b = $\emoji{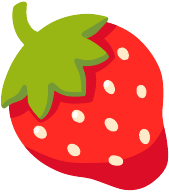}};
    \draw[<->] (A) -- node[inner sep=1pt, pos=0.5, above left] {\small $d(x,a)$} (X);
    \draw[<->] (X) -- node[inner sep=1pt, pos=0.5, above right] {\small $d(x,b)$} (B);
    \node[anchor=north west, align=center] at (0, 2.5) {
        ABX \on fruit, \by color, \across size
    };
\end{tikzpicture}
    \setlength{\belowcaptionskip}{-10pt}
    \caption{The ABX discrimination task. The test is successful if $d(x, a) < d(x, b)$. The \on attribute is the same for $a$ and $x$, and different for $b$. \by is the same for $a$, $b$ and $x$. \across is the same for $a$ and $b$, and different for $x$.}
    \label{fig:abx}
\end{figure}
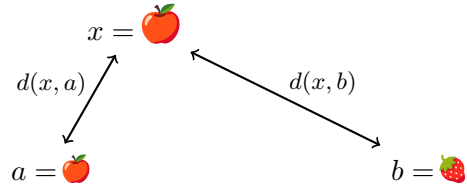

This shift has prompted deeper analysis of learned representations using various methods, from targeted probes for speaker and phonetic information \citep{liu23j_interspeech} to probe-free approaches for analyzing word \citep{pasad-etal-2024-self} or phonetic-level information \citep{wells22_interspeech}.

The ABX discriminability task \citep{schatz13_interspeech,schatz2016abxdiscriminability}
is inspired by match-to-sample tasks used in human psychophysics, and measures the discriminability between two categories. It is a zero-resource evaluation metric that does not rely on training an additional probe. It measures what is directly extractable from the representations. It is dimensionality-agnostic and works with dense or discrete representations.

\begin{figure*}[ht]
    \centering
    \includegraphics[width=\linewidth]{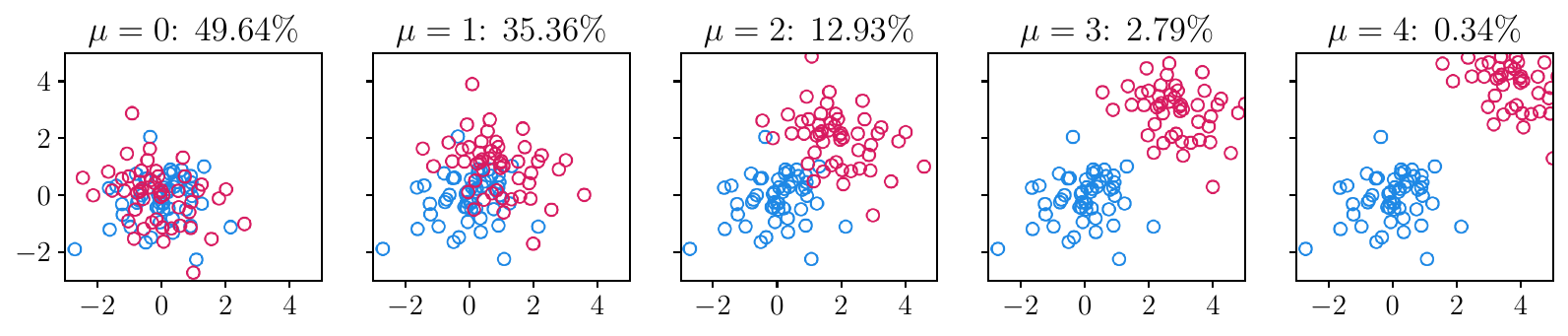}
    \caption{ABX error rate between samples from two 2D Gaussians with increasing shift. The Gaussians follow $\mathcal{N}(\mathbf{0}, I)$ and $\mathcal{N}(\bm{\mu}, I)$, with $\bm{\mu} = (\mu, \mu)$.}
    \label{fig:gaussians}
\end{figure*}

This metric has been central to the ZeroSpeech challenges \citep{9888095} for the "acoustic unit discovery" task, and has become a standard evaluation tool for SSL speech models. It has also proven particularly useful in spoken language modeling from raw audio, where speech representations are discretized and treated as pseudo-text, enabling language model training directly from speech. Studies have demonstrated that ABX discrimination scores strongly correlate with a downstream language models' ability to generate coherent speech \citep{lakhotia-etal-2021-generative}, though this can come at the cost of speech reconstruction quality \citep{poli-etal-2024-improving,défossez2024moshispeechtextfoundationmodel}. Additionally, the ABX task has been extensively employed in studies simulating human early phonetic learning \citep{doi:10.1073/pnas.2001844118,lavechin2025simulating,poli2024modeling,blandon2025simulating}, or comparing speech models and human perception abilities \citep{millet-dunbar-2022-self}. While ABX has been mostly used to evaluate speech representations, it is a generic framework that can be applied to other domains of representation learning.

We present fastabx, a library to compute ABX discriminability efficiently. It provides a simple interface that can be adapted to any specification of the ABX, and to any input modality. We believe that this tool would benefit all communities around representation learning, and open up new ways to inspect the representations of self-supervised models. 

\section{Background}

The ABX discriminability, illustrated in \cref{fig:abx}, measures how well categories of interest are separated in the representation space by determining whether tokens from the same category are closer to each other than to those from a different category. The A, B, and X in the name ABX refer to the methodology. The discriminability of category $A$ from category $B$ is the probability that a token $x$ of category $A$ is closer to another $a \in A$ than to a token $b \in B$.  For example, to measure the discriminability of the phoneme /a/ from /e/, we construct $A$ as the set of all the instances of /a/ in our corpus and $B$ as all instances of /e/. \Cref{fig:gaussians} is an example of the ABX task where the categories to discriminate are the indices of the underlying Gaussians.

In this initial formulation, categories have a single attribute: \textit{phoneme} in our example. However, in many cases, the input signal is characterized simultaneously by multiple attributes. In speech, for instance, the signal at a given time window can be characterized both by the underlying phoneme being uttered and by additional factors such as the surrounding context (previous and following phonemes), and by speaker's identity. This additional information can be used to build rich ABX tasks that test the extent to which discriminability remains robust despite variability induced by one or several other categories.

\begin{table}
    \centering
    \setlength{\tabcolsep}{4.5pt}
    \begin{tabular}{lccc}
\toprule
Task     &  $a$ & $b$ & $x$ \\
\midrule
\on fruit & \emoji{red-apple} & \emoji{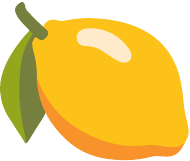} & \emoji{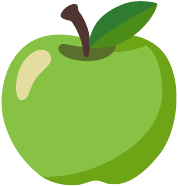} \\
\on color & \emoji{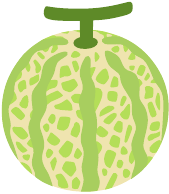}  & \emoji{red-apple} & \emoji{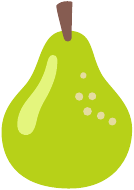} \\
\on fruit, \by color & \emoji{red-apple} & \emoji{strawberry} & \emoji{red-apple} \\
\on fruit, \by color, \across size & \emoji{green-apple} &\emoji{pear} & \scalebox{0.75}{\emoji{green-apple}} \\
\bottomrule
\end{tabular}
    \caption{Example of valid triples for various ABX tasks. Only the attributes specified in the \on, \by and \across conditions are used to build valid triples.}
    \label{tab:triples}
\end{table}

We can therefore construct an ABX task specified by three conditions, illustrated in \cref{tab:triples}. We say that we measure the ABX discriminability \on the attribute that is identical between the $A$ and $X$ categories, and that is different for $B$. We measure \by the attribute that remains the same for $A$, $B$ and $X$. Finally, when an attribute is the same for $A$ and $B$ but different for $X$, we say that the measure is \across this attribute. There can be more than one \by or \across attribute. For example, in the standard ABX task that was used in the ZeroSpeech challenges, we measure the ABX discriminability \on phoneme, \by context, and \by or \across speaker, using representations of triphones.

\begin{figure*}
    \centering
    \input{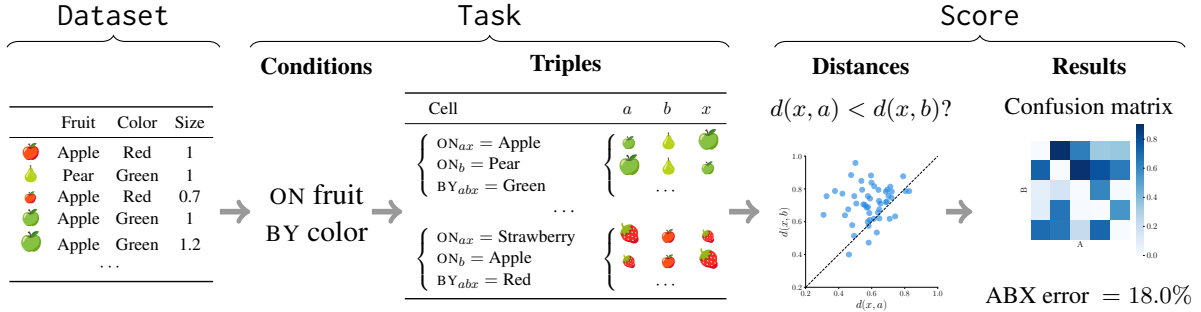}%
    \caption{The ABX discrimination task performed end-to-end. First, we build a \dataset containing the samples and their attributes. Then, the \task class precomputes all possible triples given the \on condition, and the optional \by and \across conditions. Finally, the \score class calculates $d(x, a)$ and $d(x, b)$ for every triple of every \cell of the \task, and outputs the final results.}
    \label{fig:pipeline}
\end{figure*}

We call \textit{cell} the set of triples $ \mathcal{C} = A \times B \times X$. One example of a cell in the standard phoneme ABX task is the set of all instances of /bag/ and /beg/ spoken by a given speaker. The comparison between tokens is performed using a distance $d$ on the representations of $a$, $b$ and $x$. The test is successful if the representations of $a$ and $x$ are closer than the representations of $b$ and $x$. Formally, the ABX discriminability of a cell $\mathcal{C}$ is

\[
\mathcal{D}_\mathcal{C} = \frac{1}{|\mathcal{C}|} \sum_{(a, b, x) \in \mathcal{C}} \mathds{1}_{d(a, x) < d(b, x)} + \frac{1}{2} \mathds{1}_{d(a, x) = d(b, x)}.
\]

The overall ABX discriminability, denoted by $\mathcal{D}$, is a weighted average across all cells. The weighting function is a way to balance the effects of the asymmetries between cells and the differences in cell size. What was done in the phoneme ABX task was to average first over contexts, then over the speaker identities, and finally over phonemes. Throughout the paper, we report the results in terms of the ABX error rate $1 - \mathcal{D}$.

\subsection{Previous libraries}

The first library to implement the ABX discrimination task was ABXpy\footnote{\url{https://github.com/bootphon/ABXpy}}. ABXpy could be used to build any kind of ABX task, without assuming a particular set of conditions. This was the implementation first used in the ZeroSpeech challenges \citep{versteegh2015zero,zrc2017,dunbar19_interspeech}. ABXpy was particularly slow in pre-computing the triples and building the valid cells. It is no longer maintained, and it is not compatible with recent versions of Python. The interface and the naming conventions of fastabx largely follow those of ABXpy.

The other implementation is the one distributed with Libri-Light \citep{kahn2020librilight}. Much faster than ABXpy, it is the implementation used in the ZeroSpeech 2021 challenge \citep{nguyen2020zeroresourcespeechbenchmark}. However, it had fully hardcoded the phoneme ABX task: the evaluation code would iterate over all possible contexts, then over all speakers, and then over all pairs of phonemes. This made it difficult to extend to new settings, and not suited at all for computing the ABX discriminability on anything over than speech. For example, subsequent work by \citet{hallap2023evaluating} studied the context-invariance of speech representations, by removing the context condition in the ABX task. Doing this change required rewriting a large portion of the original Libri-Light evaluation codebase.

Therefore, there was a need for a library that at the same time could provide the flexibility and generality of ABXpy, while being fast enough for quick iteration.

\section{Library overview}

The fastabx library aims to be both as fast as possible in forming triples and calculating the distances, and flexible enough to use any configuration of \on, \by, and \across condition for this ABX task. The library aims to be clear and minimal to make its maintenance easy, and the code readable and quick to understand. It should be easy to incorporate different components into one’s personal code, and use it beyond just a black box. It is distributed as a Python package\footnote{\url{https://pypi.org/project/fastabx}}, bundling a PyTorch C++ / CUDA extension, under a MIT license. It depends only on PyTorch \citep{paszke2019pytorch} and on the Polars dataframe library \footnote{\url{https://pola.rs}}.

\subsection{Interfaces}
First, the library provides one function that can be used out of the box: \py{zerospeech_abx}. This function computes the triphone or phoneme ABX, the same way as what has been done in past ZeroSpeech challenges. It is also available through a command line interface. The description of the dataset is given by an `item' file, a format introduced in ABXpy. It is a tabular format, with columns specifying the timestamps of triphones, the speaker information, the file name, etc.

The full evaluation pipeline is illustrated in \cref{fig:pipeline}. The main interface of the library consists of three classes: \dataset, \task, and \score.

The \dataset is a simple wrapper to the underlying corpus: it is made of labels and of a way to access the representations. We provide several class methods to create a \dataset from arrays, CSV files, or using an item file and a function to extract representations. This class can be easily extended to new use cases, on new types of data. It is the interface between the specific problem at hand, and the ABX evaluation itself.

\begin{python}[Dataset](fontsize=\noexpand\footnotesize)
from fastabx import Dataset

dataset = Dataset.from_item(
  item, # Path to the item file
  root, # Path to the pre-extracted features
  frequency, # Feature frequency (in Hz)
  feature_maker=torch.load,
  extension=".pt",
)       
\end{python}

The ABX \task is built given a \dataset and the \on, \by and \across conditions. It efficiently precomputes all cell specifications using the lazy operations of the Polars library. The \task is an iterable where each member is an instance of a \cell. A \cell contains all instances of $a$, $b$ and $x$ that satisfy the specified conditions for a particular value.

\begin{python}[Task](fontsize=\noexpand\footnotesize)
from fastabx import Task

task = Task(
  dataset,
  on="#phone",
  by=["prev-phone", "next-phone", "speaker"],
)

print(len(task))
# 117927
print(task[0])
# Cell(
#        ON(#phone_ax = AO, #phone_b = IH)
#        BY(speaker_abx = 6295)
#        BY(next-phone_abx = NG)
#        BY(prev-phone_abx = L)
# )
\end{python}

To control the size and number of cells, a \task can be instantiated with an additional \subsampler. The \subsampler implements the two subsampling methods done in Libri-Light. First, it can cap the number of $a$, $b$ and $x$ independently in each cell. Second, when \across conditions are specified, it can limit the number of distinct values that $x$ can take for the \on attribute.

Once the task is built, the actual evaluation is conducted using the \score class. A \score is instantiated with the \task and the name of a distance (such as `angular', `euclidean', etc.). After the scores of each \cell have been computed, they can be aggregated using the \py{collapse} method. The user can either obtain a final score by weighting according to cell size, or they can aggregate by averaging across subsequent attributes.

\begin{python}[Score](fontsize=\noexpand\footnotesize)
from fastabx import Score

score = Score(task, "angular")
abx_error_rate = score.collapse(
  levels=[
    ("prev-phone", "next-phone"),
    "speaker",
  ]
)
print(abx_error_rate)
# 0.033783210627340875
\end{python}

\subsection{Dynamic Time Warping on GPU}
In speech, the representation of a particular token has a time dimension. To compare the representations of different utterances, we need to either pool along the time domain or to find an alignment between the them. This alignment is computed using Dynamic Time Warping (DTW).
In previous libraries, the DTW algorithm was implemented in Cython. We re-implemented it as a PyTorch C++ extension with both CPU and CUDA backends.
On CPU, the computation is parallelized across the triples inside a cell using OpenMP. On CUDA devices, we implement two levels of parallelism: the first across triples using CUDA block dimension (similar to the CPU implementation), and the second within the DTW computation itself, using CUDA threads.

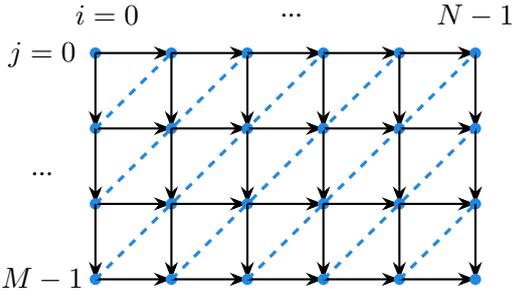
\begin{figure}
    \centering
    \usetikzlibrary{arrows.meta}%
\definecolor{myblue}{HTML}{1E88E5}%
\begin{tikzpicture}[
    y=-1cm, 
    node/.style={circle, fill=myblue, inner sep=1.5pt},
    arr/.style={-{Stealth}, line width=0.8pt}
]
\def\N{5}
\def\M{3}
\foreach \i in {0,...,\N} {
    \foreach \j in {0,...,\M} {
        \node[node] at (\i,\j) {};
        \ifnum\i<\N
            \draw[arr] (\i,\j) -- (\i+1,\j);
        \fi
        \ifnum\j<\M
            \draw[arr] (\i,\j) -- (\i,\j+1);
        \fi
    }
}
\foreach \d in {0,...,\numexpr\N+\M\relax} {
    \pgfmathsetmacro{\startx}{max(0, \d-\M)}
    \pgfmathsetmacro{\starty}{min(\d, \M)}
    \pgfmathsetmacro{\endx}{min(\d, \N)}
    \pgfmathsetmacro{\endy}{max(0, \d-\N)}
    \pgfmathparse{(\startx != \endx) || (\starty != \endy) ? 1 : 0}
    \ifnum\pgfmathresult=1
        \draw[myblue, dashed, line width=1.2pt] (\startx,\starty) -- (\endx,\endy);
    \fi
}
\draw[draw=none] (\N,-0.5) -| (-0.7,\M)
  node[pos=0]{$N-1$}
  node[pos=0.2125]{...}
  node[pos=0.425]{$i=0$}
  node[pos=0.575]{$j=0$}
  node[pos=0.8]{...}
  node[pos=1]{$M-1$};
\end{tikzpicture}
    \caption{Simple wavefront parallelism. The arrows represent the dependencies of the DTW. The elements of each diagonal are all processed in parallel using CUDA threads (dotted line).}
    \label{fig:wavefront}
\end{figure}

The DTW cost $c$ between two vectors of length $N$ and $M$ is computed via dynamic programming. The cost at step $(i, j)$ is $c_{i,j} = d(i, j) + \min \left( c_{i - 1, j}, c_{i, j - 1}, c_{i - 1, j - 1} \right)$, with  $1 \leq i \leq N -1$ and $1 \leq j \leq M -1$. Each diagonal of the DTW matrix depends of the previous two diagonals. Because of these inherent dependencies, the DTW cannot be parallelized in a straightforward manner. We have implemented a simple wavefront parallelism approach, illustrated in \cref{fig:wavefront}, where computations along each diagonal can proceed in parallel. The previous two diagonals are stored in CUDA shared memory for fast access. While the number of active threads increases and decreases throughout computation, creating a suboptimal thread utilization pattern, we found this approach to be sufficient for our use case. Further optimization could be achieved by employing tiled computation cells (see \citet{10.1145/2751205.2751243} for an overview of the different approaches to this problem), but this was not critical for our implementation given the relatively small time dimension of speech representations of triphones.

\subsection{Comparison with existing libraries}

\begin{table}[ht]
\centering
\begin{tabular}{lcc}
    \toprule
    Library & Generic & Speed \\
    \midrule
    ABXpy & \greencheckmark & 2 hr 12 min 25 s \\
    Libri-Light & \redcross  & 4 min 08 s  \\
    fastabx & \greencheckmark & 2 min 02 s \\
    \bottomrule
\end{tabular}
\caption{Comparison between the three libraries. The speed is measured by the wall time to perform the ABX task on LibriSpeech dev-clean \by speaker, without subsampling.}
\label{tab:speed}
\end{table}

We report in \cref{tab:speed} the elapsed real time to perform the ABX discrimination task. The evaluation setting follows ZeroSpeech 2021: computing the ABX error rate \on phoneme, \by context, and \by, on LibriSpeech \citep{panayotov2015librispeech} dev-clean subset, with representations of triphones. The benchmark was done on a machine with one Nvidia Tesla V100 16GB GPU and one Intel Cascade Lake 6248 processor with 20 cores.

Results from ABXpy are exactly replicated by fastabx, but this is not the case for Libri-Light by default. Indeed, during the development of fastabx, we found that the Libri-Light implementation of how features were sliced was wrong. The representations were always one frame too short at the end. See \cref{sec:slicing} for more details. We leave an option in fastabx to replicate the behavior of Libri-Light, by setting an environment variable. With this variable set, fastabx exactly replicates Libri-Light.

\section{Examples}
In addition to the core functionality of the library, we provide a set of illustrative examples. These analyses were only possible thanks to the generic aspect of fastabx, and the ease of access to the fine details of the scores, not just the aggregated one.

\subsection{Phoneme ABX and speaker ABX}

\begin{figure}
    \centering
    \includegraphics[width=\linewidth]{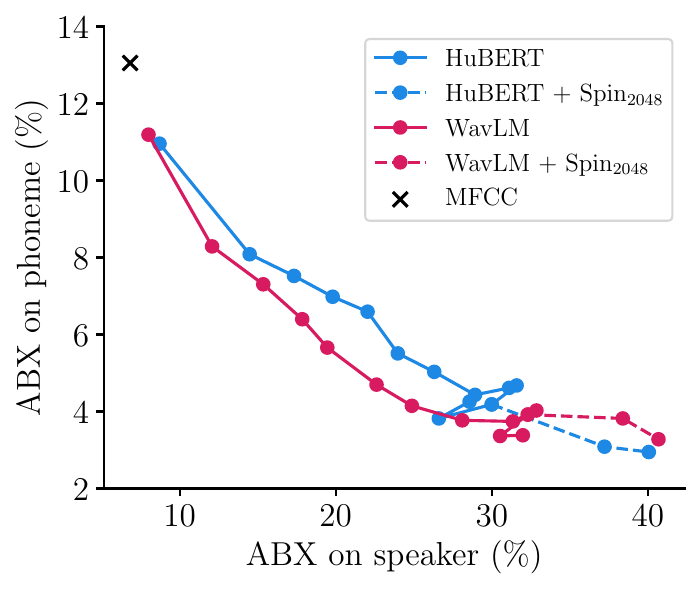}
    \caption{Trajectories along layers of the ABX on phoneme as a function of the ABX on speaker. The first layer is in the top left of the plot. Only the last two layers have been finetuned for the models with Spin: the trajectories resume from the layer 10 of the base model.}
    \label{fig:triphone}
\end{figure}

\VerbatimFootnotes
Speech SSL models are reported to learn pseudo-phonetic units, as evidenced by strong ABX scores and high mutual information with actual phoneme alignments \citep{hsu2021hubert}. However, these representations remain vulnerable to variations in acoustic environments and speaker characteristics. Recent approaches addressing these limitations include WavLM \citep{chen2022wavlm}, which improves noise robustness, and Spin \citep{chang23_interspeech}, which enhances speaker invariance—both building upon the HuBERT base architecture. WavLM incorporates an additional pretraining iteration with a novel denoising objective. Spin implements a speaker-invariant swapped prediction loss, fine-tuning only the final two layers. To evaluate these approaches, we computed two complementary ABX scores across LibriSpeech dev-clean and dev-other: phoneme discrimination (ABX \on phoneme, \by speaker and \by context) and speaker discrimination (ABX \on speaker, \by phoneme and \by context). \Cref{fig:triphone} represents the relation between those two scores, for all layers of HuBERT base and WavLM base, with or without Spin. The ABX using MFCC features\footnote{Computed with \verb|torchaudio.compliance.kaldi.mfcc|.} was added for reference. For both WavLM and HuBERT, adding Spin fine-tuning worsens the ABX on speaker, which is exactly what is expected from the method as it optimizes for speaker invariance. 

\subsection{Correlation with articulatory features}

\begin{figure}
    \centering
    \includegraphics[width=\linewidth]{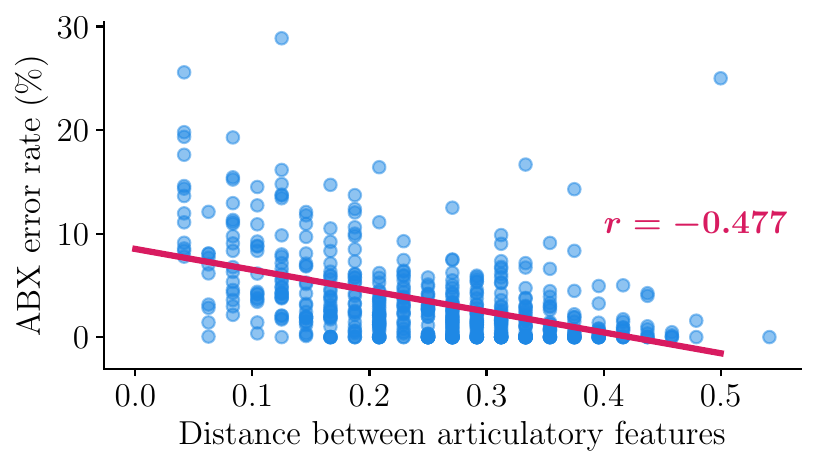}
    \caption{Correlation between the ABX error rate of a phonetic contrast and the distance between the articulatory features. The contrasts involving diphthongs were not considered.}
    \label{fig:correlation}
\end{figure}

The ABX framework also enables fine-grained analysis that provides deeper insights into the learned representations. To understand how errors are distributed across different phonetic contrasts, we examined their correlation with underlying articulatory features in \cref{fig:correlation}. We first averaged error rates across contexts and speakers, then symmetrized the scores to obtain a single error rate for each unordered phoneme contrast. Using the PanPhon library \citep{mortensen-etal-2016-panphon}, we calculated the distance between the articulatory features of each phoneme pair. The distance between an unspecified value and a specified value was set to $0.5$, and the distance between two features with opposite values to $1$.

\section{Conclusion}
We introduced fastabx, a high-performance Python package for generic ABX discrimination tasks. The library provides a comprehensive framework for evaluating self-supervised and unsupervised model representations without requiring downstream probe training. It aims to make ABX testing more accessible and practical for the entire representation learning landscape, from speech processing to a wide range of other domains.

For future work, we envision improving the performance of the CUDA backend of the DTW, integrating new subsampling methods, and introducing new interfaces to create instances of a \dataset adapted to other modalities than speech.

\section*{Acknowledgments}
The authors thank Mathieu Bernard for the preliminary work on the design of a new ABX codebase.

This work was performed using HPC resources from GENCI-IDRIS (Grant 2023-AD011014368) and was supported in part by the Agence Nationale pour la Recherche (ANR-17-EURE-0017 Frontcog, ANR10-IDEX-0001-02 PSL*, ANR19-P3IA-0001 PRAIRIE 3IA Institute) and a grant from CIFAR (Learning in Machines and Brains) awarded to E.D. in his EHESS capacity. M. P. acknowledges Ph.D. funding from Agence de l’Innovation de Défense.
\bibliography{anthology,custom}

\appendix
\section{Slicing features}
\label{sec:slicing}
To compute phoneme or triphone based ABX, we need phone-level alignments. We compute the representations using the full audio file, and we then slice to only get the frames that correspond to the unit of interest. Since the frames are downsampled, there is a decision to make on exactly which frame to keep and which to remove. Let $\onset$, $\offset$ the times of start and end of the triphone or phoneme considered provided by the alignments, with $\onset < \offset$. Let $\Delta t$ the constant time step between consecutive features, 20 ms for example. In practice, $\Delta t$ is set by the downsampling ratio of the model, due to the strides in the convolutions. We follow ABXpy, and define the set of frames indices to select $I$ as

\begin{equation}
    I = \left\{ i \in \mathbb{N} \mid \onset \leq t_i \leq \offset \right\},
\end{equation}

with $t_i = \frac{\Delta t}{2} + \Delta t \times i$  the discrete times associated to the features.

We have, for any $i \in \mathbb{N}$,
\begin{equation}
    i \in I \Leftrightarrow \begin{cases}
        i \geq  \frac{\onset}{\Delta t} - \frac{1}{2} \\
        i \leq \frac{\offset}{\Delta t} - \frac{1}{2}
        \end{cases}.
\end{equation}

Therefore, the first and last indices (both included) are:

\begin{align}
    i_\text{start} & = \min(I) = \left\lceil \frac{\onset}{\Delta t} - \frac{1}{2} \right\rceil, \\
    i_\text{end} & = \max(I) = \left\lfloor \frac{\offset}{\Delta t} - \frac{1}{2} \right\rfloor.
\end{align}

Defining $I$ and $t_i$ in this way is a choice in itself---one that is rather conservative, favoring fewer frames over more when a decision must be made. In Libri-Light, the same convention was followed, and the same $i_\text{start}$ and $i_\text{end}$ were used. However, because the features were sliced by doing \verb|features[i_start:i_end]| instead of \verb|features[i_start:i_end+1]|, the last included index was $i_\text{end} - 1 = \left\lfloor \frac{\offset}{\Delta t} - \frac{1}{2} \right\rfloor - 1$. This is especially problematic for features with a large $\Delta t$, like 40 or 80ms.
\end{document}